\documentclass[letterpaper, 10 pt, conference]{ieeeconf}  

\IEEEoverridecommandlockouts                              

\usepackage{graphics} 
\usepackage{amsmath} 
\usepackage{amssymb}  

\usepackage{microtype}
\usepackage{graphicx}
\usepackage{subfigure}
\usepackage{booktabs}
\usepackage{amsfonts}
\usepackage{amsbsy}
\usepackage{float}
\usepackage{cite}

\newfloat{figtab}{htb}{fgtb}
\makeatletter
\newcommand\figcaption{\def\@captype{figure}\caption}
\newcommand\tabcaption{\def\@captype{table}\caption}
\makeatother
\usepackage{makecell}
\usepackage{algorithm}
\usepackage{algorithmic}
\usepackage{etoolbox}
\usepackage{caption}
\makeatletter
\patchcmd{\@makecaption}
{\scshape}
{}
{}
{}
\makeatother

\title{\LARGE \bf
CRIL: Continual Robot Imitation Learning \\via Generative and Prediction Model
}

\author{Chongkai Gao$^{1}$, Haichuan Gao$^{1}$, Shangqi Guo$^{1}$, Tianren Zhang$^{1}$, and Feng Chen$^{1}$$^{2}$$^{3}$
\thanks{$^{1}$Department of Automation, Tsinghua University, Beijing,
	100084, China. 
        {\tt\small \{gck20,ghc18,zhang-tr19\}@mails.tsinghua.}
        {\tt\small edu.cn, shangqi\_guo@foxmail.com,chenfeng@mails.}
        {\tt\small tsinghua.edu.cn} }%
\thanks{$^{2}$Beijing Innovation Center for Future Chip, Beijing 100086}%
\thanks{$^{3}$LSBDPA Beijing Key Laboratory, Beijing,
	100084. China.}%
}

\begin{document}

\maketitle
\thispagestyle{empty}
\pagestyle{empty}

\begin{abstract}

Imitation learning (IL) algorithms have shown promising results for robots to learn skills from expert demonstrations. However, they need multi-task demonstrations to be provided at once for acquiring diverse skills, which is difficult in real world. In this work we study how to realize continual imitation learning ability that empowers robots to continually learn new tasks one by one, thus reducing the burden of multi-task IL and accelerating the process of new task learning at the same time. We propose a novel trajectory generation model that employs both a generative adversarial network and a dynamics-aware prediction model to generate pseudo trajectories from all learned tasks in the new task learning process. Our experiments on both simulation and real-world manipulation tasks demonstrate the effectiveness of our method. 

\end{abstract}

\section{INTRODUCTION}

Intelligent robots nowadays are expected to develop diverse skills to operate well in the real world. Although imitation learning (IL) has shown the power for robots to acquire skills from expert demonstrations\cite{reinforcementandimitationlearningfordiversevisuomotorskills},\cite{oneshothierarchicalimitationlearning}, it suffers when the robot is required to learn numerous tasks simultaneously, since it is unrealistic to provide robots with all necessary demonstrations in advance \cite{rethinkingcontinuallearningforautonomous}, and it is also quite a challenge for IL algorithms today to handle with various expert guidance\cite{analgorithmperspectiveofimitationlearning}. One way to tackle this problem is to perform \textit{continual learning} in IL tasks to let the robot continuously update its policy using demonstrations from potentially unlimited new tasks over its lifetime while avoiding the abrupt degradation of previously learned skills\cite{marginalreplayvsconditionalreplay}. Continual IL can relieve the burden of collecting demonstrations and enables IL algorithms to learn only one task at one time, thus serving as a practical way to develop versatile robots in the real world.

A number of previous works have been studying this problem. Most of them aim to use training data from learned tasks in the training process of a new task to maintain previous knowledge\cite{policydistillation,onlinemultimodalimitationleanring,lifelonginversereinforcementlearning,memoryefficientexperiencereplay}. However, to acquire real data of learned tasks, they either need a large buffer to store all data of learned tasks, or assume the robot can go back to previous environments to be trained in them again, while keeping environments or raw training data of past tasks unchanged is basically infeasible  in real-world scenarios\cite{rethinkingcontinuallearningforautonomous}. Additionally, privacy issues are also need to be considered for mass storage.

\begin{figure}[t]
	\centering
	\includegraphics[width=0.95\linewidth]{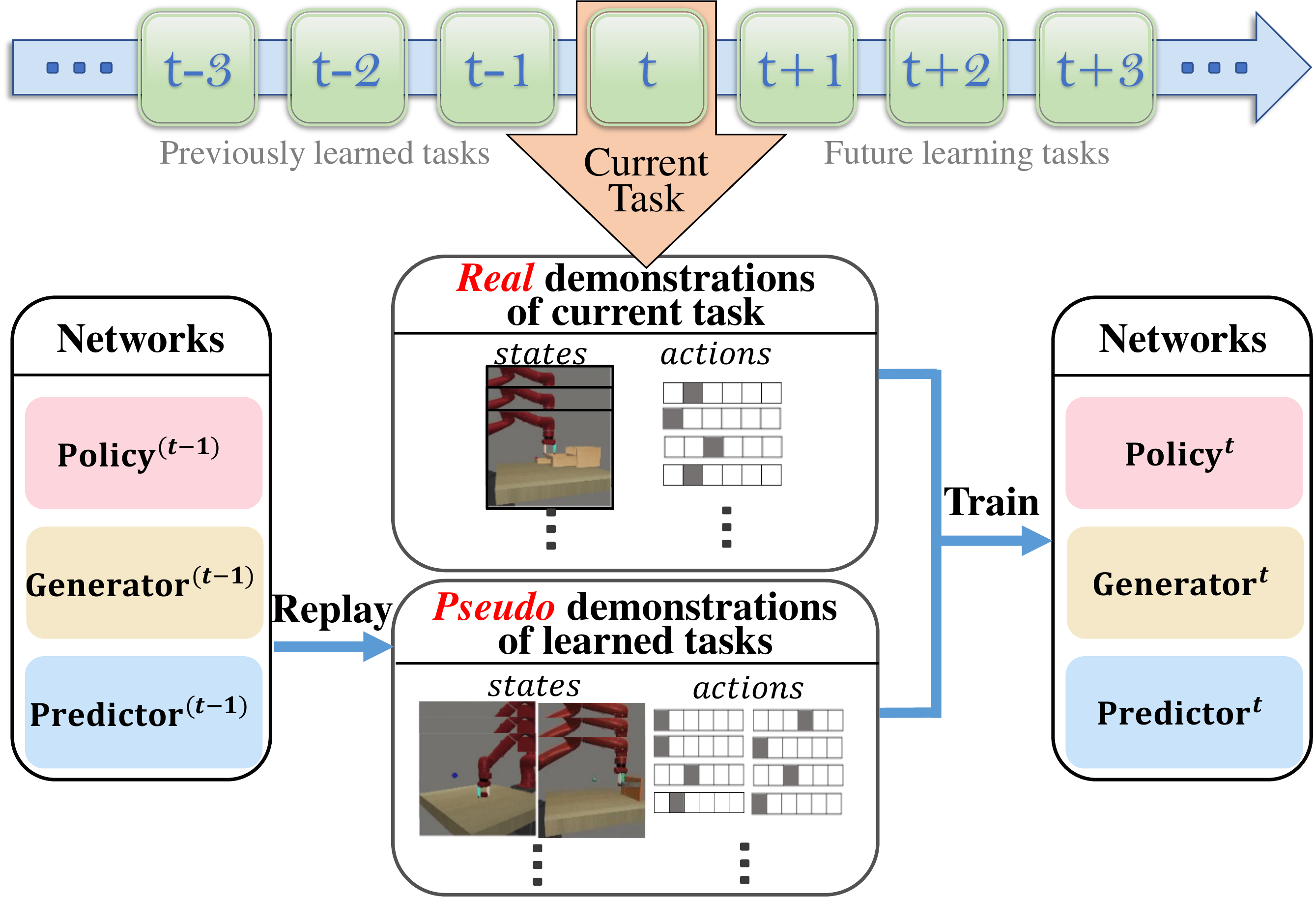} 
	\caption{The paradigm of CRIL. It replays pseudo data of learned tasks and interleaves them with real data of new task to update its networks.}
	\label{fig:intro}
\end{figure}

This problem comes from that these approaches all need to obtain \textit{real data} of learned tasks for the training of the new task. However, this is not essentially required in continual IL. Recently, in the machine learning domain, a prevailing idea called deep generative replay (DGR) \cite{continuallearningwithdeepgenerativereplay} aims to produce \textit{pseudo data} of learned tasks in the training of the new task. DGR employs a generative model to memorize the data distribution\cite{generativememoryforlifelonglearning} of learned tasks and generates samples from the learned distribution when learning on new tasks and interleaves them
with real data of new task to update its models, therefore avoiding the need to obtain real data.

The key challenge to apply DGR to continual robot IL lies in how to generate satisfactory pseudo data of learned tasks. The problem is that robot demonstrations are trajectories, i.e., time-series. The long time-series data property
hugely increases the dimension of the target distribution space, and most video generation models today\cite{timegan},\cite{mocogan} struggle to capture the temporal dependence contained in the trajectory data. These problems make it difficult for a single generative model to capture the data space of the whole trajectory\cite{stochasticvideogeneration}. However, unlike ordinary videos that only consist of state information, robot trajectories here contain rich action information provided by experts. This unique information can be leveraged to help the generative model capture the transition dynamics in the trajectories by predicting the next frame according to the previous frame and action, thus avoiding generating the whole video at once.

Following this idea, in this work we present CRIL, a DGR-based method for continual robot IL tasks designed for trajectory generation. Concretely, we use generative adversarial networks to generate only the first frame of each trajectory from learned tasks, and use an action-conditioned video prediction network to predict subsequent frames of the same trajectory based on generated states and actions from DGR policy. As illustrated in Fig. \ref{fig:model}, our method jointly generates states and actions in the replaying stage and explicitly captures the stepwise transition dynamics in the demonstration trajectories. CRIL alleviates the training complexity of the generator by reducing the search space of the GAN model from the original large video space to a much smaller single image space and transfers most of the training complexity to the prediction model. The prediction model is supervised learned by abundant frames in the provided demonstrations, thus can be well trained and the quality of generated trajectories can be guaranteed.

The main contribution of this work is that we are the first to successfully apply the deep generative replay idea to continual robot IL tasks and give out a specialized implementation for robot trajectory generation. In our experiments, we show that our approach can achieve continual learning ability in both simulation and real-world tasks. We evaluate the proposed model on diverse manipulation tasks both in meta-world\cite{metaworld} tasks, and several manipulation tasks using a Jaco2 robot arm in the real world. Both experiment results show the superiority of our method.

\section{RELATED WORK}

\subsection{Continual Imitation Learning for Robotics}

Achieving continual learning ability is an open question in
the robot IL domain\cite{analgorithmperspectiveofimitationlearning}, \cite{continuallearningforrobots}. It requires robots to learn from demonstrations in a sequence of tasks while avoiding catastrophic forgetting\cite{catastrophic}, in which the model's performance on previous tasks abruptly degrades when trained on a new task. Dynamically expanding the policy network\cite{progressive} and elastic weight consolidation\cite{ewc} are feasible ways to go, but the first way consumes lots of memory space, and the second way does not perform as good as replay-based methods that replay and retrain robots with previous data\cite{rethinkingcontinuallearningforautonomous},\cite{continuallearningforrobots}. 

For replay-based methods, recently \cite{onlinemultimodalimitationleanring} and \cite{lifelonginversereinforcementlearning} borrow the idea of ELLA\cite{ELLA} to store and update a set of reward function basis learned from demonstrations by inverse reinforcement learning with the help of stored data. \cite{policydistillation} and \cite{continualreallife} use policy distillation to distill the knowledge of previously learned policies to a new policy. \cite{memoryefficientexperiencereplay} and \cite{unicorn} use rehearsal to store all the training data of previous tasks in a buffer and then use them in the training of new tasks. \cite{continualstaterepresentationlearningforreinforcementlearning} and \cite{incrementalrobotlearningofnewobjects} use state representation learning to build a general perception model for objects and environments, and then use the representation to facilitate the new task learning process, such as curiosity-driven exploration\cite{curiosity}. However, these methods have different assumptions that make them impractical and lack scalability in real-world robot tasks: having a large storage space\cite{lifelonginversereinforcementlearning},\cite{memoryefficientexperiencereplay},\cite{unicorn}, having access to previous environments\cite{policydistillation},\cite{onlinemultimodalimitationleanring},\cite{continualreallife}, or similar skills can be adopted in different tasks\cite{continualstaterepresentationlearningforreinforcementlearning},\cite{incrementalrobotlearningofnewobjects}.

\subsection{Deep Generative Replay}

Deep generative replay\cite{continuallearningwithdeepgenerativereplay} (DGR) is a continual learning approach that alleviates catastrophic forgetting by leveraging a generative model to memorize the data distribution of learned tasks and generate samples from it in the training process of a new task\cite{generativememoryforlifelonglearning},\cite{continuallearningforrobots}. In each task training procedure, the generator is first used to generate data of learned tasks, then the policy network would label these replayed data, finally the hybrid data of learned and new tasks are used to update both policy and generator networks. DGR has been widely used to achieve continual learning ability in different domains\cite{deepgenerativedual}\cite{generativemodelsfrom}\cite{continualstaterepresentationlearningforreinforcementlearning}. The generative model can be original GAN network, WGAN-GP, conditional GAN, or variational autoencoder\cite{measuring}\cite{continuallearningwithdeepgenerativereplay}\cite{marginalreplayvsconditionalreplay}\cite{generativemodelsfrom}. Note the choice of different GAN variants is not the main point of the article. A detailed comparison of their performances for continual learning tasks can be found in \cite{marginalreplayvsconditionalreplay}. In this work we choose WGAN-GP\cite{wgangp}.

\subsection{Video Generation and Prediction}

Video generation aims to generate realistic videos based on various inputs, such as random noise\cite{tgan} or text\cite{textgan}. It can be used for DGR to replay full trajectories. Recent works usually employ RNN networks as both the  generator and discriminator of the GAN network to realize  time-series generation\cite{mocogan},\cite{vgan}. On the other hand, video prediction\cite{videopredictionmse},\cite{stochasticvideogeneration} aims to predict subsequent frames in a video given the previous frames. The difference between these two problems is that the input of video generation problems is usually only a vector drawn from some latent space\cite{mocogan}, while the input of video prediction is previous frames. The latter one can usually produce images with better quality, and according to the results of \cite{mocogan}, directly generating whole trajectories using common video generation approaches can not get enough image quality to support continual robot IL tasks, which is also verified in our experiments.

\begin{figure*}[t]
	\centering
	\includegraphics[width=0.9\linewidth]{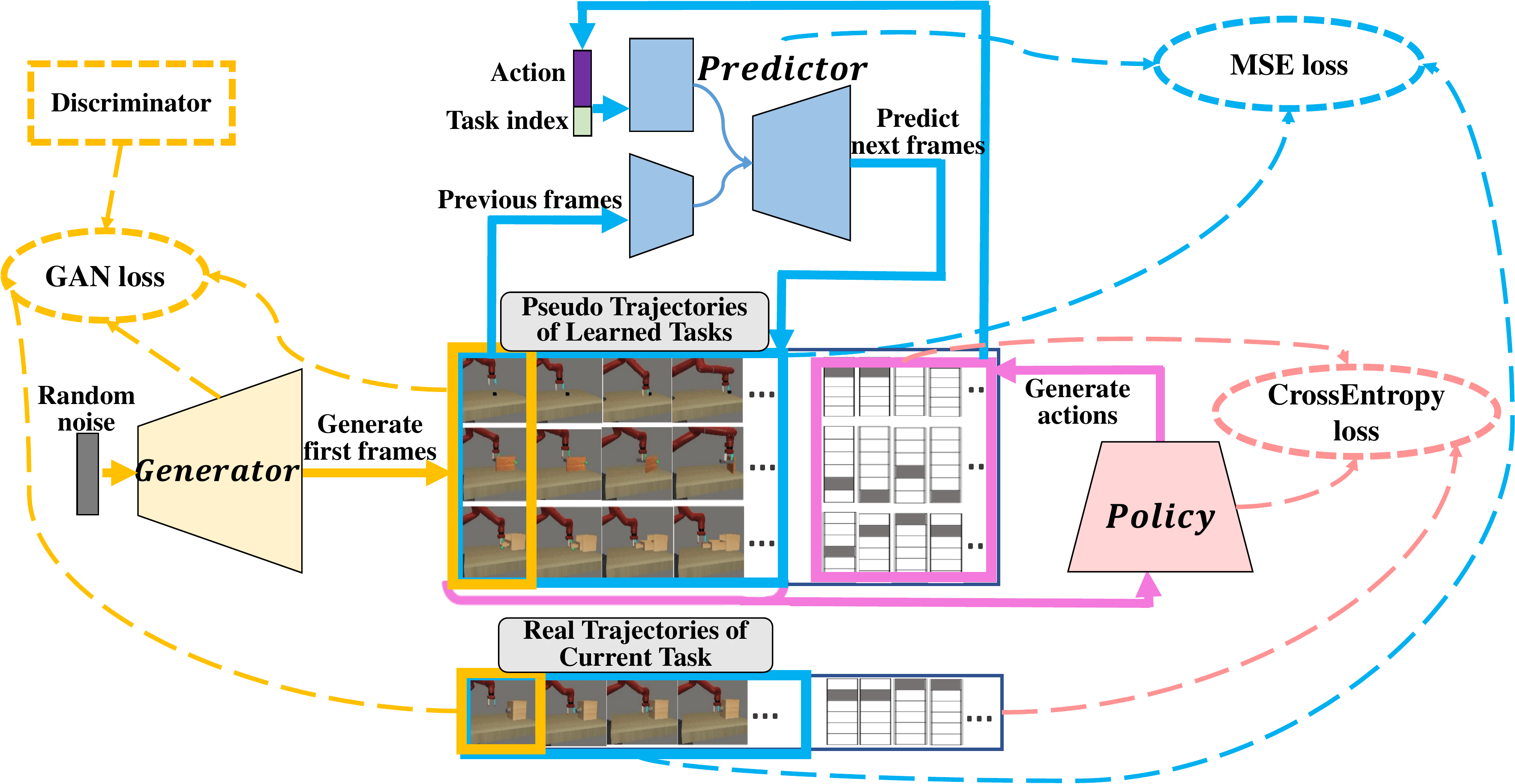} 
	\caption{Network Architecture. Solid lines represent the process of replaying pseudo data and dashed lines represent the loss for training. There are three main modules in CRIL: The generator generates first frames of different trajectories, and the policy and the predictor will iteratively generate actions and subsequent frames. These modules will be trained according to their own loss functions respectively, which are introduced in Section 3 and Section 4.}
	\label{fig:model}
\end{figure*}

\section{METHOD}
\label{section3}

In this section, we analyze how to apply DGR to continual robot IL tasks. We first give the formulation of the continual robot IL problem, and then analyze it and propose CRIL to tackle this problem.

\subsection{Background: Imitation Learning}

Robot IL algorithms aims to learn a behavior
policy $ \pi_{\theta} $ parameterized by $ \theta $ through mimicking a set of expert demonstrations\cite{reinforcementandimitationlearningfordiversevisuomotorskills},\cite{analgorithmperspectiveofimitationlearning}. Demonstrations are usually provided as a dataset of state-action pairs $ \mathcal{D}=\left\{(s_i, a_i)\right\}_{i=1\dots N } $ that belong to different trajectories $ \tau_i $ with length $ N $. These data come from one or different tasks$ \mathcal{T}_i $, which is an MDP without the reward function and the discounting factor and can be represented as a tuple: $ \left\langle \mathcal{S},\mathcal{A},T,\rho_0\right\rangle  $ with state-space $ \mathcal{S} $, action-space $ \mathcal{A} $, transition dynamics $ T: \mathcal{S}\times\mathcal{A}\times\mathcal{S}\rightarrow[0,1] $ and the initial state distribution $ \rho_0 $. Additionally, we denote the state-action distribution and state distribution of a policy by $ \rho^\pi(s,a) $ and $ \rho^\pi(s) $. There are mainly two kinds of methods for IL: behavior cloning (BC) and inverse reinforcement learning (IRL). Note in this work we focus on the continual learning ability of IL, and our method is essentially not limited to any specific IL algorithms. We choose BC since performing IRL is quite time-consuming on real-world robots.

\subsection{The Continual Robot Imitation Learning Problem}

In contrast to learning all demonstrations at one time, continual IL focuses on how to learn from demonstrations that come from online sequential tasks. Formally, based on the loss function of BC, we give out the following definition of continual IL:

\textbf{Definition 1} \textit{Continual imitation learning (CIL) aims to update policy $ \pi_{\theta} $ with a sequence of tasks $ \mathcal{T}^{(1)}, \mathcal{T}^{(2)}\dots, \mathcal{T}^{(N_{max})} $ where $ N_{max} $ is the total number of tasks. In the training of task $ \mathcal{T}^{(t)} $, it minimizes the following objective provided with state-action pairs that come from expert distribution $ \rho_{\text{exp}}^{(t)} $:}
\begin{equation}\label{lil}
\begin{split}
\mathcal{L}_{CIL}^{(t)}(\theta) =-\sum_{i=1}^{t}\lambda^{(i)}\mathbb{E}_{s^{(t)},a^{(t )}\sim\rho_{\text{exp}}^{(t)}}[\log \pi_\theta(a^{(t)} \mid s^{(t)})],
\end{split}
\end{equation}
\textit{where $ \lambda^{(i)}= 1/t,  i=1,2,\dots,t$ indicates all tasks are equally important.}

This definition reveals the biggest difficulty of optimizing the continual IL problem: in the training process of task $ \mathcal{T}^{(t)} $, we only have access to demonstrations $ \mathcal{D}^{(t)} $ from $ \rho_{exp}^{(t)} $, while we need to optimize an objective that is evaluated by data from all tasks $ \mathcal{D}^{(1,2,\dots,t)} $ that sampled from $ \rho_{exp}^{(1,2,\dots,t)} $. Thus the key problem is how to acquire $ \mathcal{D}^{(1,2,\dots,t)} $ during the training of $ \mathcal{T}^{(t)} $.

\subsection{DGR for Continual Robot Imitation Learning}

In order to get over the above difficulty, DGR generates pseudo data of learned tasks by a generative model to fill these data vacancies. Thus there are two networks in DGR for continual IL process: the generator $ G^{(t)}_\psi $ parameterized by $ \psi $ and the policy network $ \pi_\theta^{(t)} $. The full training procedure is illustrated in Fig. \ref{fig:intro} and Fig. \ref{fig:model}. At the start of training a new task, the generator would generate \textit{states} of learned tasks from its own distribution. The current policy would then label actions for these pseudo states, and then the pseudo states and actions would be mixed  together with real data of the new task to train the policy through a certain IL algorithm. Finally, the generator will be trained to fit the mixed state distribution of generated pseudo states and the real states from new task demonstrations. This mixed distribution can approximately represent the real state distribution of all learned tasks so far.

\begin{algorithm}[t]
	\caption{Continual Robot Imitation Learning}
	\label{alg:example}
	\begin{algorithmic}
		\STATE {\bfseries Initialize:} Task index $ t=0 $, policy network $ \pi_{\theta}^{(0)} $, image generation network $ G_\psi^{(0)} $,  prediction network $ P_\phi^{(0)} $, learning rates $ \lambda_\theta $, $ \lambda_\psi $ and $ \lambda_\phi $.
		
		\STATE {\bfseries While} haveNewTask() {\bfseries do}
		
		\STATE \quad $ t = t + 1 $
		
		\STATE \quad Get demonstration data $ \mathcal{D}^{(t)} $ that consist of $ m $ trajectories from new task $ \mathcal{T}^{(t)} $
		
		\STATE \quad Initialize a trajectory buffer $ B^{(t)} $ and put $ \mathcal{D}^{(t)} $ into $ B^{(t)} $
		
		\STATE \quad {\bfseries for} $ i=1,t $ {\bfseries do}
		
		\STATE \quad \quad Use $ G_\psi^{(t-1)} $ to generate $ m $ first images for $ m $ different trajectories $ \left\{\tau^{(i)}\right\} $ of task $ \mathcal{T}^{(i)} $;
		
		\STATE \quad \quad Use $ P_\phi^{(t-1)} $ and $ \pi_{\theta}^{(t-1)} $ in each trajectory $ \tau^{(i)} $ to generate the full trajectory based on first frames;
		
		\STATE \quad \quad Collect generated $ \left\{\tau^{(i)}\right\} $ and put them into $ B^{(t)} $;
		
		\STATE \quad {\bfseries end for}
		
		\STATE \quad Update networks using $ B^{(t)} $:
		
		\STATE \quad \quad $ \theta^{(t)}  \leftarrow \theta^{(t-1)} - \lambda_\theta \nabla_\theta \mathcal{L}^{(t-1)}_\pi(\theta)$

		\STATE \quad \quad $ \psi^{(t)}  \leftarrow \psi^{(t-1)} - \lambda_\psi \nabla_\psi \mathcal{L}^{(t-1)}_G(\psi)$

		\STATE \quad \quad $ \phi^{(t)}  \leftarrow \phi^{(t-1)} - \lambda_\phi \nabla_\phi \mathcal{L}^{(t-1)}_P(\phi)$

		\STATE \quad Delete buffer $ B^{(t)} $
		
		\STATE {\bfseries end while}
		
	\end{algorithmic}

\end{algorithm}

Formally, this procedure involves two independent processes. For training $ G^{(t)} $, a GAN loss is employed to capture the mixed  distribution of pseudo states and real states:

\begin{equation}\label{lossg}
\begin{split}
\mathcal{L}^{(t)}_G(\psi)&=\mathbb{E}_{s \sim p_{\text{mixed}}(s)}[\log D^{(t)}(s)] \\&\qquad \qquad+\mathbb{E}_{z}[\log (1-D^{(t)}(G_\psi^{(t)}(z)))],
\end{split}
\end{equation}
where $ p_{\text{mixed}}(s) $ is the mixed distribution of real and pseudo
states, and $ z \sim \mathcal{N}(0,1) $ is randomly sampled from a unit
Gaussian distribution. A supervised learning loss is used to train $ \pi_{\theta}^{(t)} $ to capture the action distribution conditioned on states:
 
\begin{equation}\label{policy}
\begin{split}
\mathcal{L}^{(t)}_\pi(\theta)=\mathbb{E}_{(s, a) \sim p_{\text{mixed}(s,a)}} [\log \pi_\theta(a^{(t)} \mid s^{(t)})].
\end{split}
\end{equation}
where $ p_{\text{mixed}}(s,a) $ is the mixed distribution of real and pseudo
state-action pairs. A key problem here is that what kind of generation model we should choose for $ G_\psi^{(t)} $. For
robot demonstration data, the generator needs to deal with
time-series trajectories. One direct way is to use an image generator $ G_\psi^{(t)} $ to generate independent states randomly, without considering the integrity of generated trajectories. This method treats all states of learned tasks as a disordered set, and could directly borrow any commonly used GAN models that perform image-level generation. Another way is to generate a full trajectory at one time, using a time-series GAN model like in \cite{mocogan}. We name these two ways as Original DGR and Trajectory DGR for convenience and the following comparison, as shown in Fig. \ref{fig:originaldgr} and \ref{fig:trajectorydgr}.

However, these two methods cannot perform well in our problem settings. The reason is that they all employ a single GAN network to capture the whole state space, which is much larger than common continual learning tasks, and therefore much harder for GAN networks to capture. Essentially, the problem comes from that they split the generation of states and actions, thus put the entire complexity of video generation to the generator. This comes from the traditional application scenarios of DGR, which are usually supervised learning tasks like image classification. In those tasks, there is no correlation between different states (pictures to be classified), and actions (image labels) have no effect on adjacent states. However, in most robot imitation learning tasks, two adjacent frames do have strong dynamics correlation conditioned on the action information, which can
be explicitly leveraged to substantially reduce the generation
difficulty.

\subsection{CRIL: Continual Robot Imitation Learning}

In order to integrate action generation into the state generation process, we need to consider the state-action joint distribution, rather than only the state distribution for the generator. We first note that the probabilistic distribution of a complete trajectory $ \tau $ consisting of both states and actions can be decomposed as follows:

\begin{figure}[t]
	\centering
	\subfigure[Generate disordered states]{
		\begin{minipage}[t]{0.42\linewidth}
			\centering
			\includegraphics[width=1\linewidth]{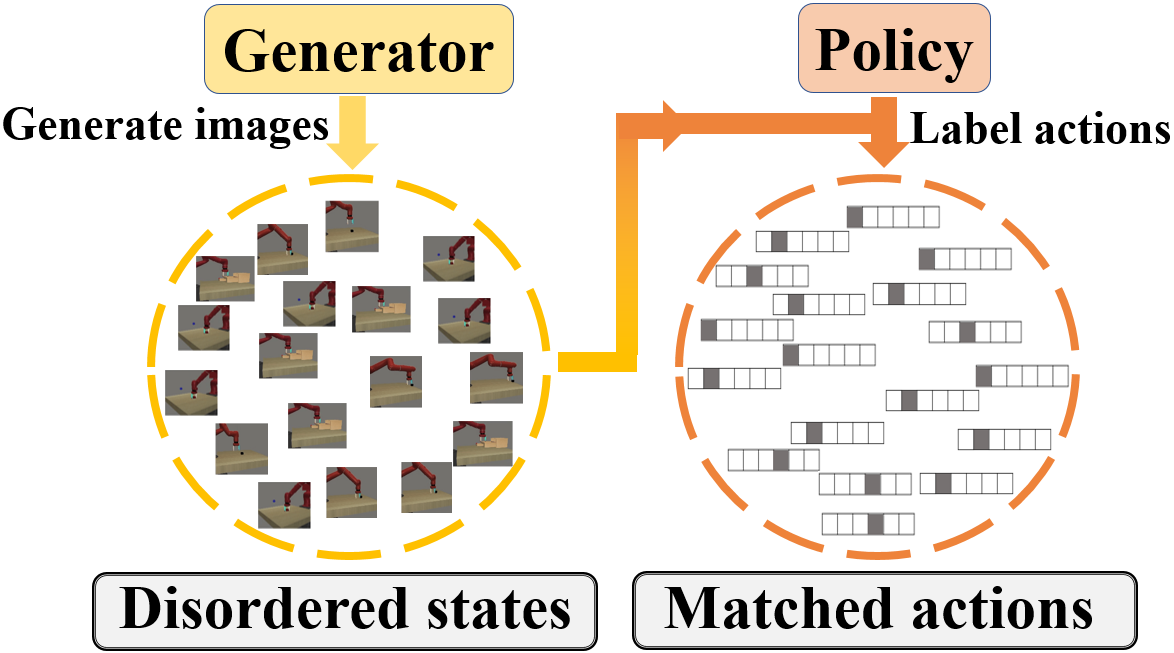}
			\label{fig:originaldgr}
		\end{minipage}
	}
	\subfigure[Generate full trajectories]{
		\begin{minipage}[t]{0.42\linewidth}
			\centering
			\includegraphics[width=1\linewidth]{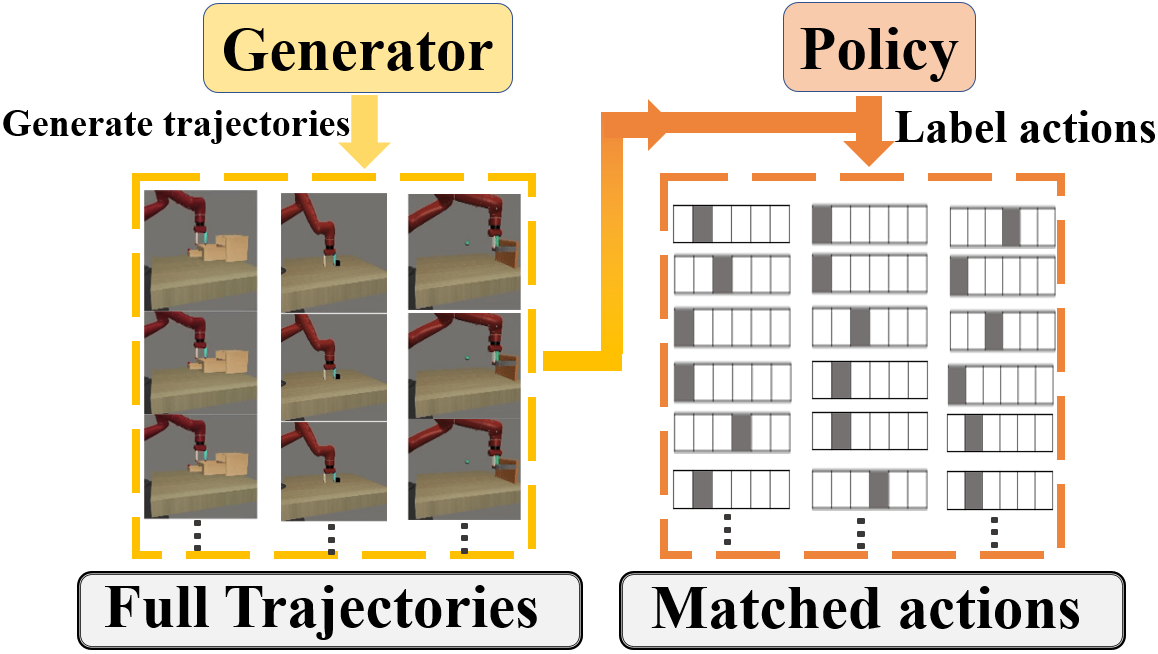}
			\label{fig:trajectorydgr}
		\end{minipage}
	}
	
	\subfigure[Generate first frames and predict subsequent frames]{
		\begin{minipage}[t]{0.9\linewidth}
			\centering
			\includegraphics[width=0.9\linewidth]{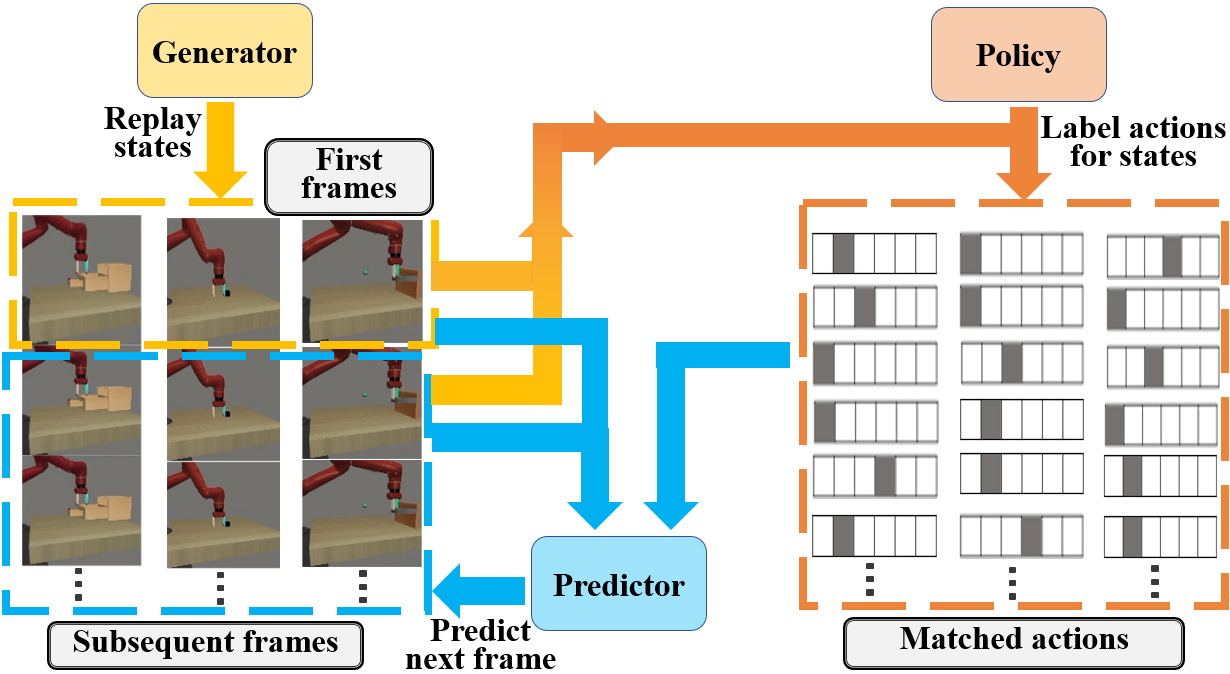}
			\label{fig:CRIL}
		\end{minipage}
	}
	\caption{Different generating strategies. }
	\label{fig:decomposition}
	\vspace{-0.2cm}
\end{figure}

\begin{equation}\label{traj}
	\rho_{\pi_{\theta}}(\tau) = \underbrace{\rho_0(s)}_{\text{first frame generator}}\prod_{i=1}^{T}\underbrace{\pi_{\theta}(a_i|s_i)}_{\text{policy}}\underbrace{p(s_{i+1}|s_i, a_i)}_{\text{next frame predictor}},
\end{equation}
where $ T $ is the number of time steps in this trajectory. This decomposition indicates us that we need three models for generating a complete trajectory: a first frame generator $ G_\psi $, a policy network $ \pi_\theta $, and a next frame predictor $ P_\phi $ parameterized by $ \phi $. As shown in Fig. \ref{fig:CRIL}, for generating a new trajectory, $ G_\psi $ first generates a start frame, and then the policy and predictor iteratively spread out the subsequent frames. By adding a predictor into the DGR process, we successfully capture the transition dynamics information to facilitate the trajectory generation. Intuitively, CRIL can
be viewed as first initializing an environment, and then reproducing the expert behaviors in this environment. The whole process of CRIL is shown in Algorithm \ref{alg:example}. The loss function of $ P_\phi^{(t)} $ is a mean square error loss:
\begin{equation}\label{lossp}
\mathcal{L}^{(t)}_P(\phi)=\sum_{i=1}^{N}\sum_{j=1}^{T}(\hat{s_{ij}} -  P_\phi^{(t)}(s_{ij},a_{ij}))^2,
\end{equation}
where $ N $ is the number of trajectories and $ T $ is the number of steps in one trajectory, and $ \hat{s_{ij}} $ are the real states. In this work we use an encoder-decoder architecture as the next frame predictor. Note that like the choice of GAN architecture, the specific type of prediction model is not the main point of this paper. One can choose to use advanced models \cite{stochasticvideogeneration}\cite{deepvisualforesight} as substitutes.

Although the high-level idea of CRIL is straightforward, its optimality needs to be proved to ensure that it can achieve the same result with the standard DGR method, e.g., Trajectory DGR. We analyze the optimality in the following section.

\section{ANALYSIS OF OPTIMALITY}

In this section, we theoretically analyze the optimality of CRIL and point out it preserves the same optimal solution as Trajectory DGR. We firstly analyze the optimizing process of Trajectory DGR and then prove the optimality of CRIL.

\subsection{The Optimizing Process of Trajectory DGR}

We denote a \textit{state trajectory} in task $ \mathcal{T}^{(t)} $ that only contains \textit{states} as $ \tau_s^{(t)} $, the distribution of generated pseudo state trajectories as $ p_g({\tau_s}^{(1,2,\dots,t)}) $, and the distribution of real state trajectories from all learned tasks as $ p_r({\tau_s}^{(1,2,\dots,t)}) $. The objective of Trajectory DGR is to minimize the distance of $ p_g({\tau_s}^{(1,2,\dots,t)}) $ and $ p_r({\tau_s}^{(1,2,\dots,t)}) $. Since we use WGAN-GP in this paper, we employ \textit{Earth-Mover} distance $ W(p_g, p_r) $ as
the distance metric. We give the following lemma that reveals the mathematical form of the optimizing process in Trajectory DGR:

\noindent\textbf{Lemma 1} \textit{ In Trajectory DGR, during the training of task $ \mathcal{T}^{(t)} $, the distance between $ p_g({\tau_s}^{(1,2,\dots,t)}) $ and $ p_r({\tau_s}^{(1,2,\dots,t)}) $ is:}
\begin{equation}\label{lemma1}
W(p_g,p_r) = \mathbb{E}_{\tau_s \sim p_m^{(t)}}[D^{(t)}(\tau_s)]-\mathbb{E}_z[D^{(t)}(G_\psi^{(t)}(z))],
\end{equation}
\textit{where}
\begin{equation}\label{lemma2}
p_m^{(t)}(\tau_s) = \frac{1}{t}[p_r({\tau_s}^{(t)}) + (t-1) p_g({\tau_s}^{(1,2,\dots,t-1)})],
\end{equation}
\textit{and the gradient direction for optimizing the trajectory generator $ G_\psi^{(t)} $ is:}
\begin{equation}\label{lemma3}
	\nabla_\psi W(p_g,p_r) = -\mathbb{E}_z[\nabla_\psi D^{(t)}(G_\psi^{(t)}(z))],
\end{equation}
\textit{where $ z \sim \mathcal{N}(0,1) $ is a Gaussian random variable and $ D^{(t)} $ is the trajectory discriminator in the training of task $ \mathcal{T}^{(t)} $}.

\textit{Proof:} In the training of task $ \mathcal{T}^{(t)} $, the target distribution that $ G_\psi^{(t)} $ aims to fit is the mixed distribution $ p_m^{(t)}(\tau_s) $ of real data from task $ \mathcal{T}^{(t)} $ and the replayed pseudo data from $ p_g({\tau_s}^{(1,2,\dots,t-1)}) $. If we assume every task is equally important, we could get equation (\ref{lemma2}), and the desired $ W(p_g, p_r) $ becomes $ W(p_g, p_m) $. According to the Kantorovich-Rubenstein duality\cite{optimaltrans}, the Earth-Mover distance $ W(p_g,p_r) $ can convert to the following form:
\begin{equation}\label{lemma4}
\begin{split}
W(p_g, p_m)= &\inf _{\gamma \sim \Pi\left(p_g, p_m\right)} \mathbb{E}_{(x, y) \sim \gamma}[\|x-y\|]\\=& \frac{1}{K} \sup _{\|f\|_{L} \leq K} \mathbb{E}_{\tau_s \sim p_m}[f(\tau_s)]-\mathbb{E}_{\tau_s \sim p_g}[f(\tau_s)],
\end{split}
\end{equation}
where $ \Pi\left(p_g, p_m\right) $ denotes all possible joint contribution of $ (p_g, p_m) $, and $ f $ lies in $ \mathcal{F} = \left\{f: \tau_s \rightarrow \mathbb{R}, f \in C_b(\tau_s), \|f\|_{L} \leq K \right\} $. Since $ \tau_s $ is compact, we can construct a neural network $ D^{(t)} $ whose parameters are restricted to satisfy $ \|f\|_{L} \leq K $ to make it lies in $ \mathcal{F} $, as in \cite{wgangp}. Then we can use $ D^{(t)} $ as $ f $ in (\ref{lemma4}) get (\ref{lemma1}). For (\ref{lemma3}), as $ \mathbb{E}_z||\nabla_\psi D^{(t)}(G_\psi^{(t)}(z)|| < +\infty $ and $ \nabla_\psi D^{(t)}(G_\psi^{(t)}(z) $ is well defined almost everywhere\cite{generativememoryforlifelonglearning}, by the dominated convergence, we could exchange the integral and derivation symbol to get (\ref{lemma3}):
\begin{equation}\label{lemma5}
\begin{split}
\nabla_\psi W(p_g,p_r) &= -\nabla_\psi \mathbb{E}_z[ D^{(t)}(G_\psi^{(t)}(z))] \\&= -\mathbb{E}_z[\nabla_\psi D^{(t)}(G_\psi^{(t)}(z))].
\end{split}
\end{equation}

\subsection{Optimality of CRIL}

\begin{figure*}[t]
	\centering
	\caption* {\uppercase{table} \uppercase\expandafter{\romannumeral1}: Quantitative results of means and standard variances from 5 random seeds. Higher value indicates better performance.}
	
	\includegraphics[width=0.95\linewidth]{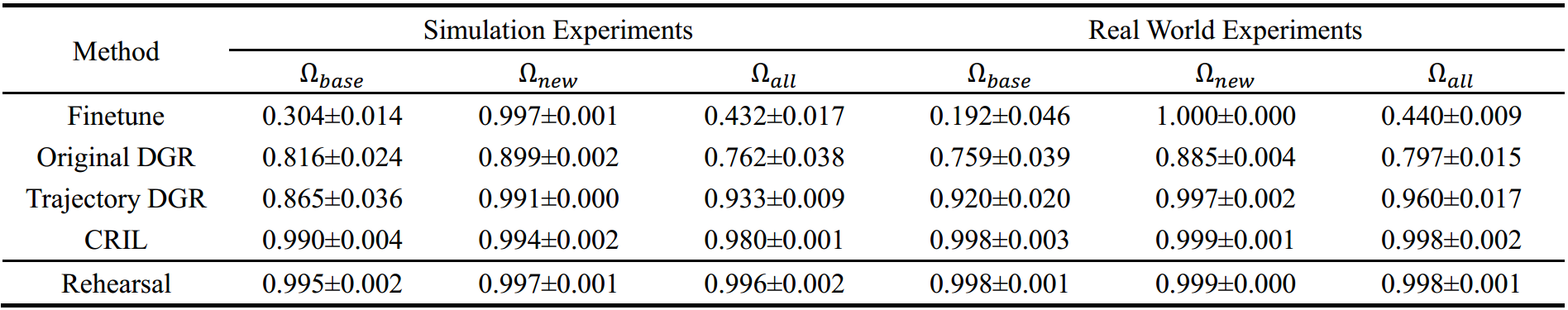} 
		\label{table:metrics}

\end{figure*}

In order to prove the optimality of CRIL, we need to prove that the loss functions (\ref{lossg}) and (\ref{lossp}) can form another eligible distance measurement that lies in $ \mathcal{F} $, and the gradient of CRIL must be equal to (\ref{lemma3}). We give out the following theorem:

\noindent\textbf{Theorem 1} \textit{ The training procedure of CRIL is equal to minimize an Earth-Mover distance between $ p_g({\tau_s}^{(1,2,\dots,t)}) $ and $ p_r({\tau_s}^{(1,2,\dots,t)}) $:}
\begin{equation}\label{theorem1}
W(p_g, p_m)= \mathbb{E}_{\tau_s \sim p_m}[\tilde{f}(\tau_s)]-\mathbb{E}_{\tau_s \sim p_g}[\tilde{f}(\tau_s)],
\end{equation}
\textit{where}
\begin{equation}\label{theorem2}
\tilde{f}(\tau_s) = D(s_0) + \frac{1}{T-1}\sum_{i=1}^{T}(s_i-\hat{s_i})^2,
\end{equation}
\textit{where $ s_{0:T} $ are the states in $ \tau_s $, and $ \hat{s_i} $ is the states from $ p_m^{(t)} $, and $ \nabla W(p_g, p_m) $ is equal to (\ref{lemma3}).}

\textit{Proof:} Since $ D^{(t)} $ here is also a generator that trained with WGAN-GP method, $ D(s_0) $ can be Lipschitz continuous just like it in Trajectory DGR . On the other hand, for $ (s_i-\hat{s_i})^2 $, through image normalization, $ s_i $ and $ \hat{s_i} $ can both lie in $ [-1,1] $, thus (\ref{theorem2}) is Lipschitz continuous. Meanwhile, we can train the image discriminator $ D $ to increase $ D(s_0) $ and train $ P_\phi $ to make $ \frac{1}{T-1}\sum_{i=1}^{T}(s_i-\hat{s_i})^2 $ to zero to get the supremacy of all eligible functions in $ \mathcal{F} $. This is exactly the max-min procedure of optimizing standard GAN networks. Thus CRIL becomes the same problem of Trajectory DGR.

The only thing left here is to prove optimizing (\ref{theorem1}) can lead to the same optimal solution of Trajectory DGR to prove CRIL can preserve the optimality. This could be proved by explaining the original gradient direction (\ref{lemma5}) is equal to the gradient of (\ref{theorem2}). We rewrite the trajectory $ \tau_s $ as a vector $ \left[s_0, s_1, \dots, s_T\right] = \left[G_\psi(z), P_\phi(s_0, a_0), \dots, P_\phi(s_{T-1}, a_{T-1})\right]$ and get the following derivation of (\ref{lemma5}):
\begin{equation}
\begin{split}
&\mathbb{E}_z[\nabla D^{(t)}(G^{(t)}(z))]\\=
&\mathbb{E}_{\tau_s}[\nabla D^{(t)}(G^{(t)}_\psi(z), \nabla P^{(t)}_\phi(s_0, a_0), \dots, P^{(t)}_\phi(s_{T-1}, a_{T-1}))] \\=
&\mathbb{E}_z[\nabla_\psi D^{(t)}(G^{(t)}_\psi(z)), 0, \dots, 0] \\ 
& +\mathbb{E}_{s_{1:T-1}}[\nabla_\phi D^{(t)}(0, P^{(t)}_\phi(s_0, a_0), \dots, P^{(t)}_\phi(s_{T-1}, a_{T-1}))] \\ =
& \mathbb{E}_z[\nabla_\psi D^{(t)}(G^{(t)}_\psi(z))]\\
&+\mathbb{E}_{s_{1:T-1}}[\nabla_\phi \sum_{t=1}^{T-1}(\hat{s_{t+1}} - P^{(t)}_\phi(s_t, a_t))],
\end{split}
\end{equation}
which equals to the gradient of (\ref{theorem1}). Thus CRIL can preserve the optimality of solution. 

To summarize, CRIL is also minimizing an Earth-Mover distance between the same two distributions in Trajectory DGR, thus preserves the optimality. As we all know, training a time-series GAN to capture the whole video data space is quite difficult. Under the premise of preserving the optimality, CRIL reduces most of the generation complexity of GAN module to make it only generate the first frames. Thus our method becomes much easier and faster to train.

\begin{figure}[t]
	\centering
	\subfigure[Simulation environments]{
		\begin{minipage}[t]{0.45\linewidth}
			\centering
			\includegraphics[width=1\linewidth]{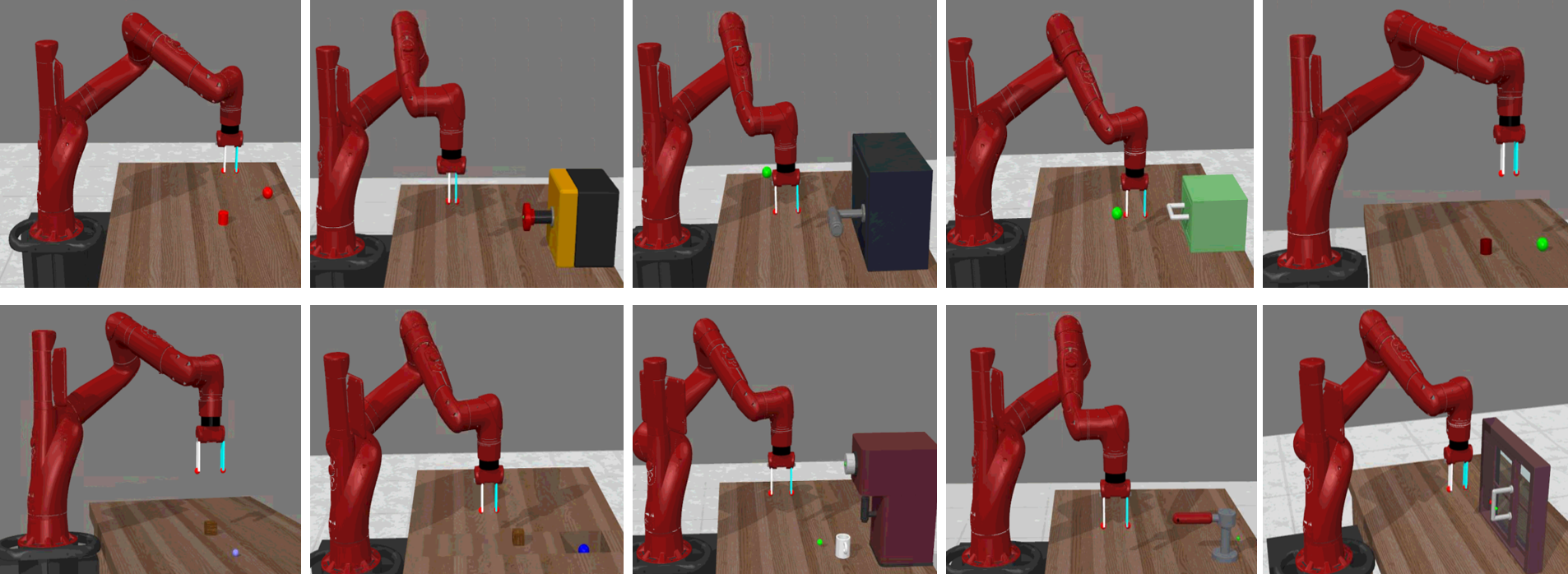}
			\label{fig:env1}
		\end{minipage}
	}
	\subfigure[Simulation topview]{
		\begin{minipage}[t]{0.45\linewidth}
			\centering
			\includegraphics[width=1\linewidth]{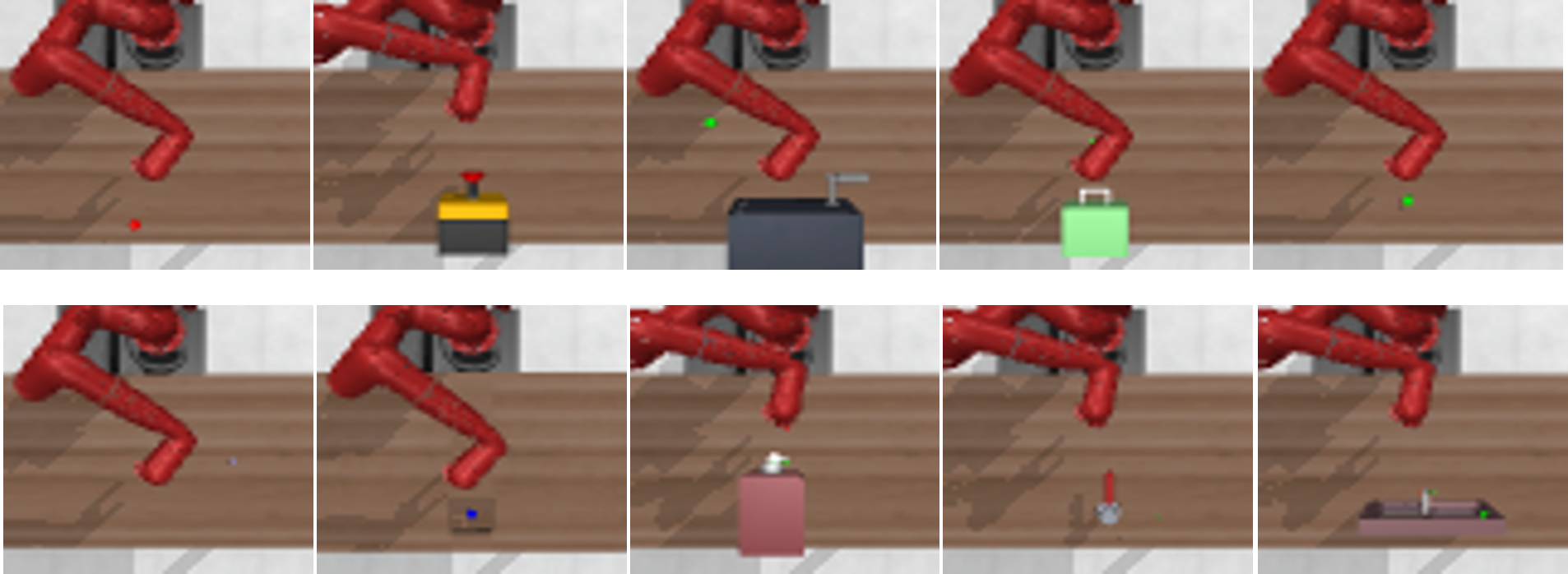}
			\label{fig:env2}
		\end{minipage}
	}
	
	\subfigure[Real world tasks]{
		\begin{minipage}[t]{0.45\linewidth}
			\centering
			\includegraphics[width=1\linewidth]{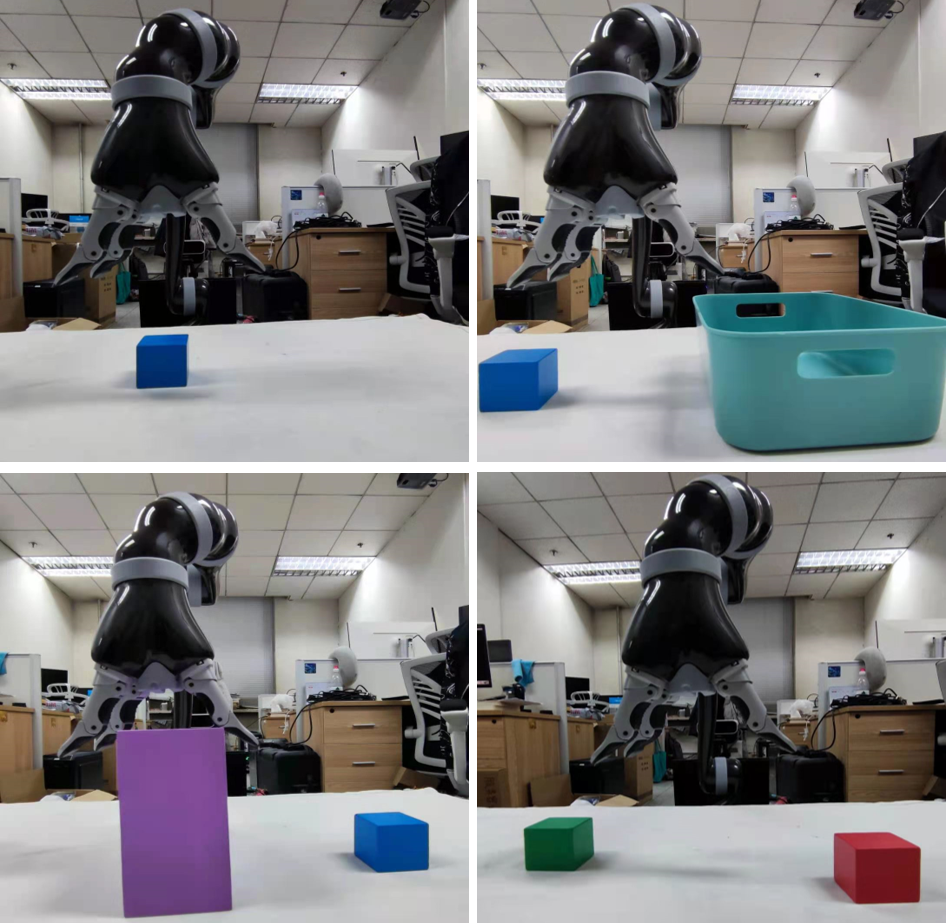}
			\label{fig:env3}
		\end{minipage}
	}
	\subfigure[Real world topview]{
		\begin{minipage}[t]{0.45\linewidth}
			\centering
			\includegraphics[width=1\linewidth]{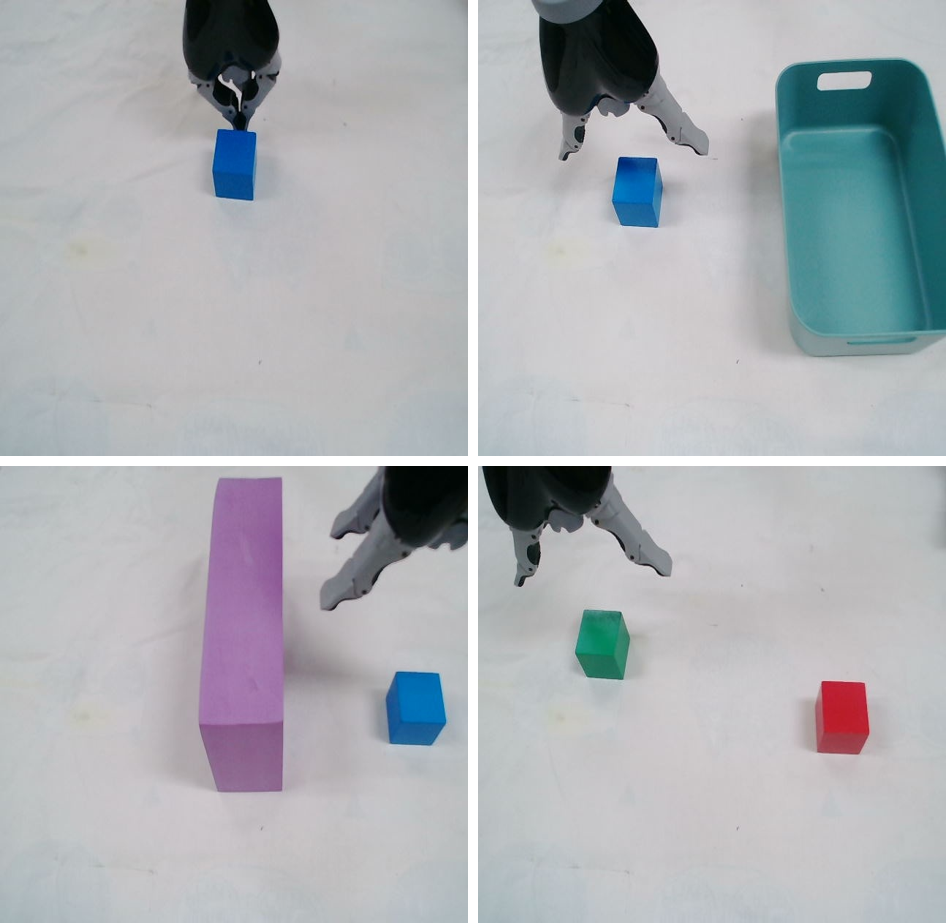}
			\label{fig:env4}
		\end{minipage}
	}
	\caption{Tasks in our experiment. Simulation tasks are: reach, button-press, door-open, drawer-open, push, sweep, sweep-into, coffee-button, faucet-open and window-open. Real world tasks are: push, place, pick-over and stack. The cameras in all environments are set at the topview position.}
	\label{fig:envs}
\end{figure}

\section{EXPERIMENTS}

Our experiments aim to answer the following questions: (1) can CRIL be successfully applied to robot continual IL problems in both simulation environments and real-world environments? (2) how is the performance of our method compared to Original DGR and Trajectory DGR? and (3) how does our proposed predictor affect the generating result? We answer (1) and (2) in the quantitative section and answer (3) using qualitative results. Codes are available at https://github.com/HeegerGao/CRIL.

\subsection{Experiment Setup}

We set up two separated groups of tasks:

\textbf{Simulation environments:} We choose ten manipulation tasks from the meta-world benchmark\cite{metaworld} as shown in Fig. \ref{fig:env1} and \ref{fig:env2}. The initial configuration of each task is randomized within certain limits. For each simulation task, we collect 100 trajectories as expert demonstrations. Each trajectory contains 23 frames to 124 frames, where each frame is a 3x64x64 RGB image from a top-view camera. The action space of robot is a discrete space that consists of seven different actions, including six actions for moving along six axes directions and a stop action that indicates the end of the task.

\begin{figure*}[h]
	\centering
	\subfigure[Images generated by Original DGR]{
		\begin{minipage}[t]{0.3\linewidth}
			\centering
			\includegraphics[width=1\linewidth]{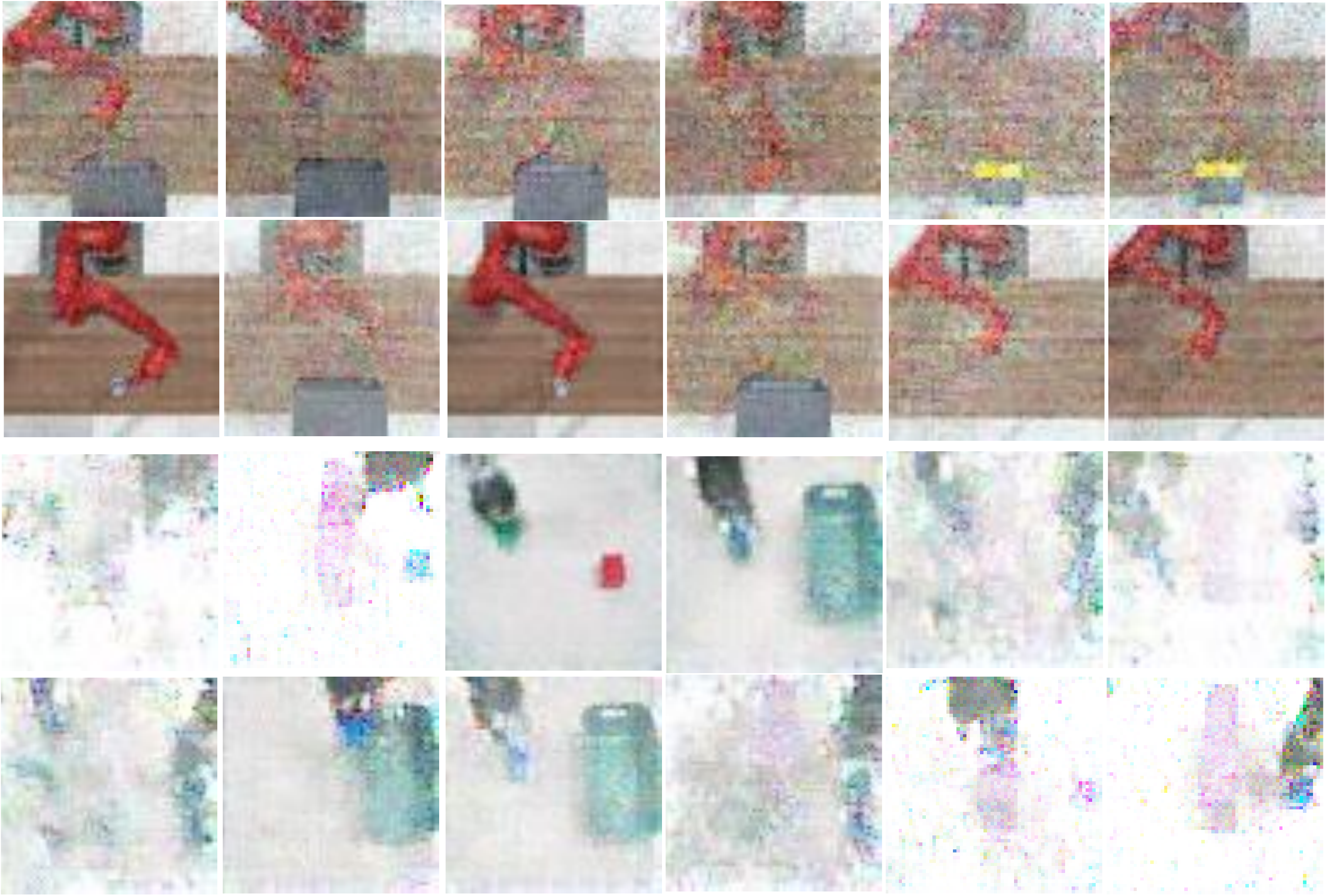}
		\end{minipage}
	}
	\subfigure[Images generated by Trajectory DGR]{
		\begin{minipage}[t]{0.3\linewidth}
			\centering
			\includegraphics[width=1\linewidth]{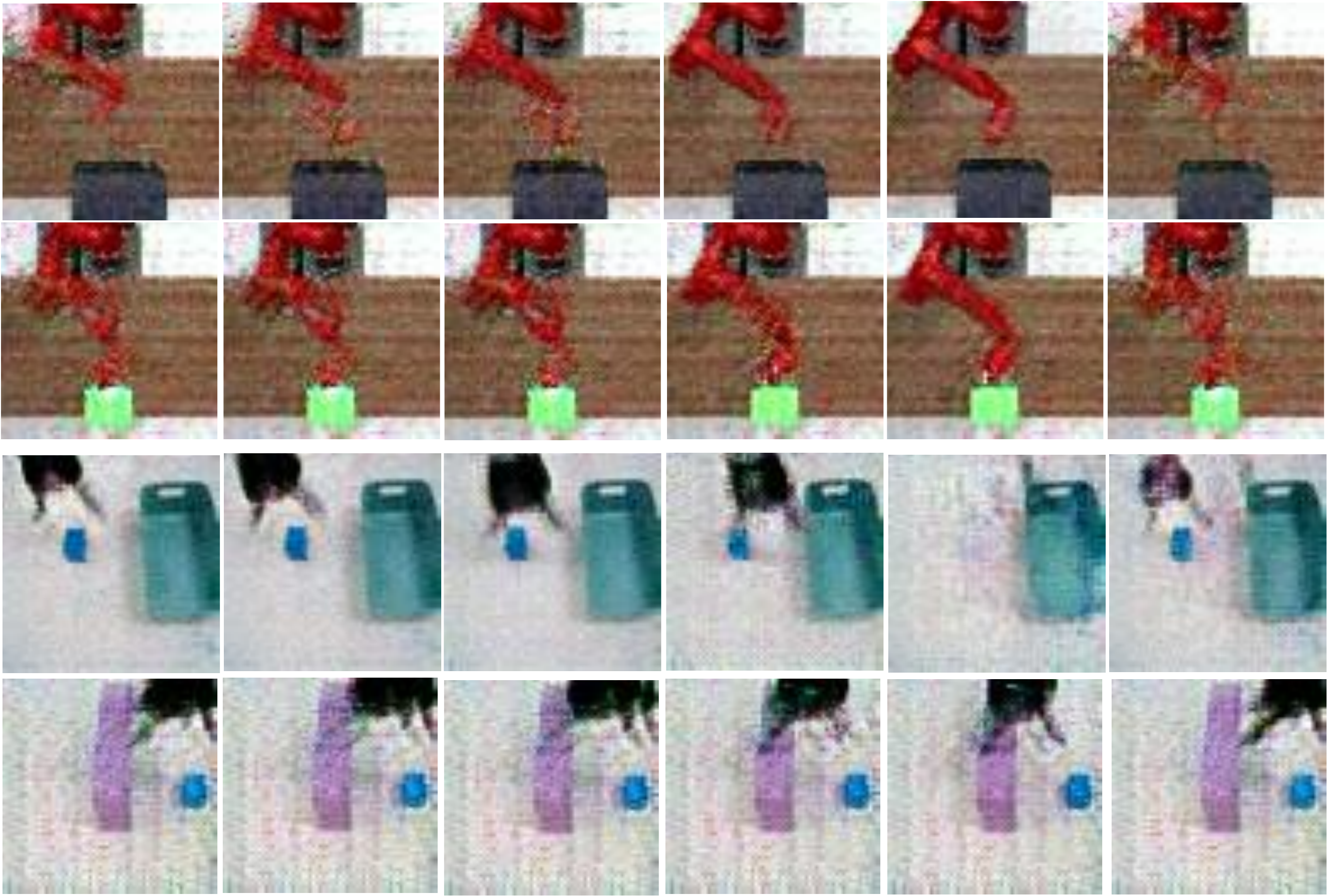}
		\end{minipage}
	}
	\subfigure[Images generated by CRIL]{
		\begin{minipage}[t]{0.32\linewidth}
			\centering
			\includegraphics[width=1\linewidth]{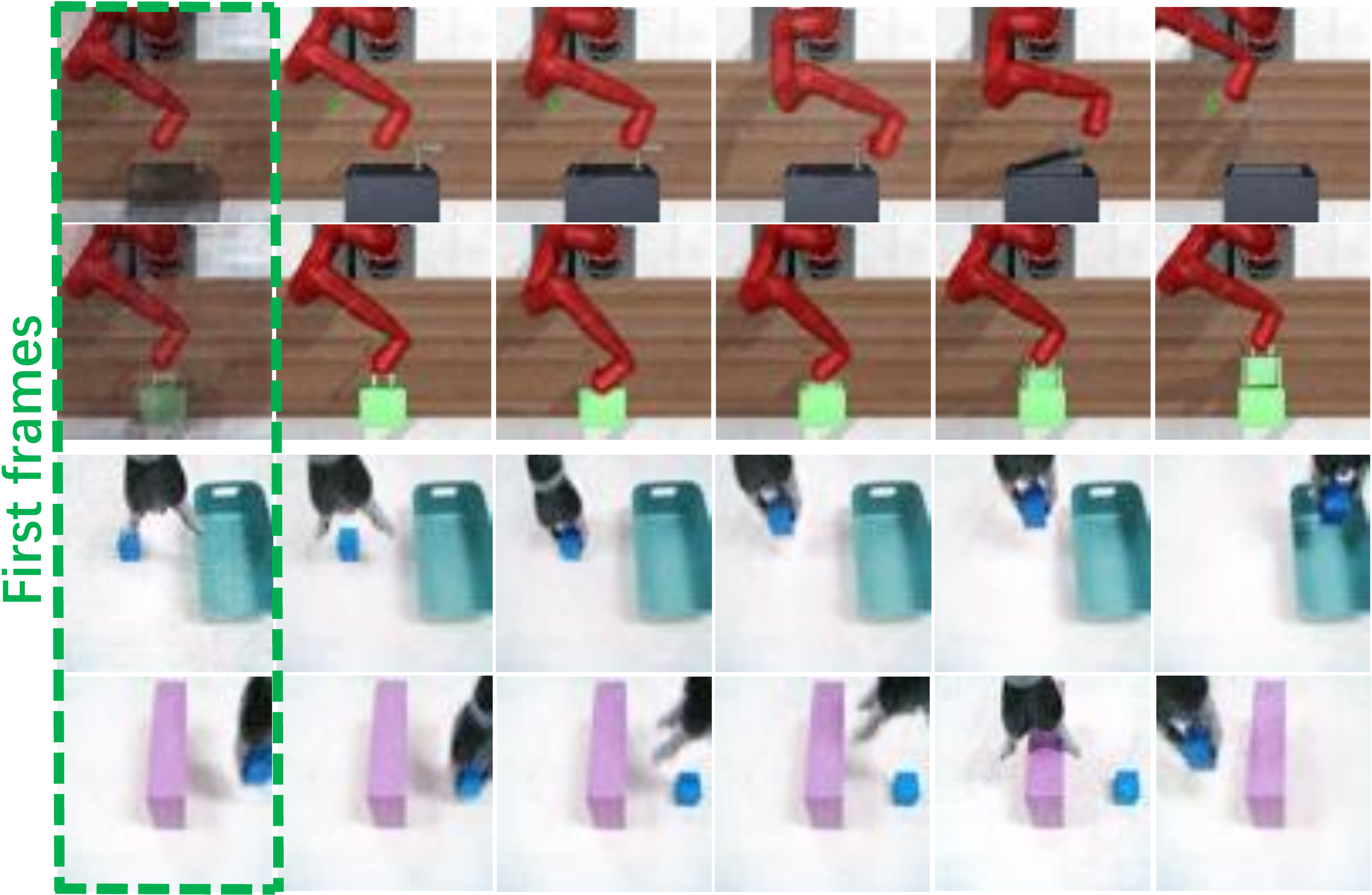}
		\end{minipage}
	}
	
	\caption{Images generated by CRIL, Original DGR and Trajectory DGR after all tasks are leaned. For CRIL and Trajectory DGR, we select four whole trajectories to show. For Original DGR, since the integrity of trajectory can not be guaranteed, we can only select separated images that come from different trajectories randomly.}
	\label{fig:generatedimages}
	\vspace{-0.3cm}
\end{figure*}

\textbf{Real-world environments:} We use a 6-DoF Kinova Jaco2 robot arm and design four manipulation tasks for experiments as shown in Fig. \ref{fig:env3} and \ref{fig:env4}. We set an RGB camera at the top view position of each task to collect image observations which are all in a 3x64x64 size. We use the keyboard to control the Cartesian position of the end effector of the robot arm to give out demonstrations. For each task, we collect 15 trajectories as expert demonstrations, which takes us about 1 minute for collecting each trajectory. The action space here is the same as in simulation environments.

\subsection{Methods and Metrics} 

We compare five different methods in the above continual IL setting. We implement Original DGR and Trajectory DGR with two prevailing generation networks respectively. In all tasks we use the \textit{success rate} as the basic indicator for evaluation:
\begin{itemize}
	\item \textbf{Finetune:} No continual learning algorithm performed, which means the policy would be updated using only new demonstrations. 
	
	\item \textbf{Rehearsal:} The real training data of learned tasks is directly replayed for policy training via a large replay buffer. This result is the upper bound of the performance of all other methods.
	
	\item \textbf{Original DGR:} The method in Fig. \ref{fig:originaldgr}. We use WGAN-GP\cite{wgangp} as our generator to generate separated images.
	
	\item \textbf{Trajectory DGR:} The method in Fig.  \ref{fig:trajectorydgr}. We borrow MoCoGAN\cite{mocogan} as our generator model. MoCoGAN decomposes motion and content in videos and achieves SOTA performance on the video generation domain.
	
	\item \textbf{CRIL:} The proposed model in this work.
		
\end{itemize}

We adopt the commonly used metrics in continual learning proposed in \cite{measuring}, which include three criteria:

\begin{equation}
	\Omega_{base}=\frac{1}{N-1} \sum_{i=2}^{N} \frac{\alpha_{base, i}}{\alpha_{ideal }},
	\vspace{-0.1cm}
\end{equation}
\begin{equation}
	\Omega_{new}=\frac{1}{N-1} \sum_{i=2}^{N} \alpha_{new, i},
	\vspace{-0.1cm}
\end{equation}
\begin{equation}
	\Omega_{all}=\frac{1}{N-1} \sum_{i=2}^{N} \frac{\alpha_{all, i}}{\alpha_{ideal }},
\end{equation}
where $ N $ is the total number of tasks, and $ \alpha_{new, i} $, $ \alpha_{base, i} $ and $ \alpha_{all, i} $ denote the test accuracy of the new task, the first task and all learned tasks respectively, and $ \alpha_{ideal} $ is used to normalize $ \Omega_{base}$ and $ \Omega_{all}$. $ \Omega_{base}, \Omega_{new} $ and $ \Omega_{all} $ measure the model's ability to remember previous knowledge, transfer  knowledge to new tasks and the performance on overall tasks. For all three metrics, a higher value indicates better performance.

\subsection{Quantitative Comparison}

\begin{figure}[t]
	\centering
	\subfigure[Simulation Results]{
		\begin{minipage}[t]{0.46\linewidth}
			\centering
			\includegraphics[width=1\linewidth]{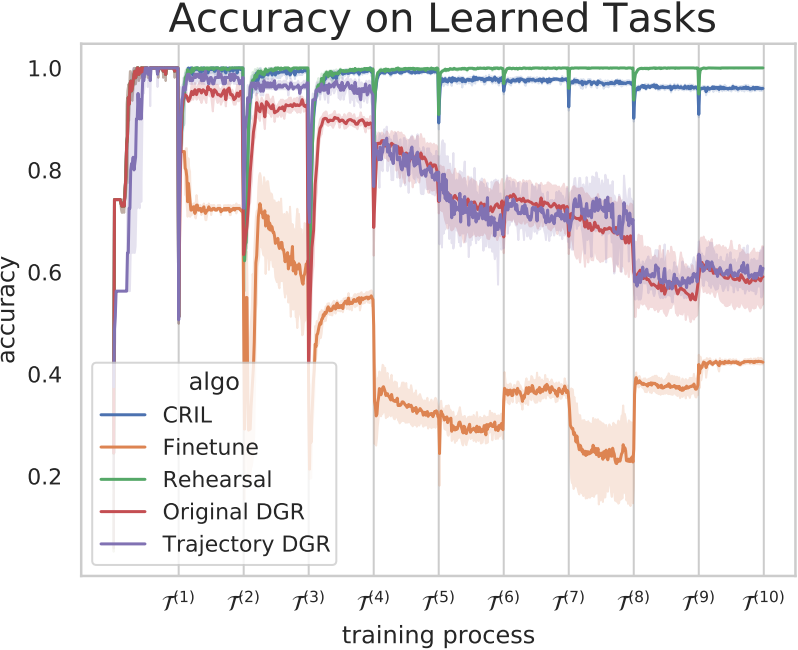}
		\end{minipage}
	}
	\subfigure[Real world Results]{
		\begin{minipage}[t]{0.46\linewidth}
			\centering
			\includegraphics[width=1\linewidth]{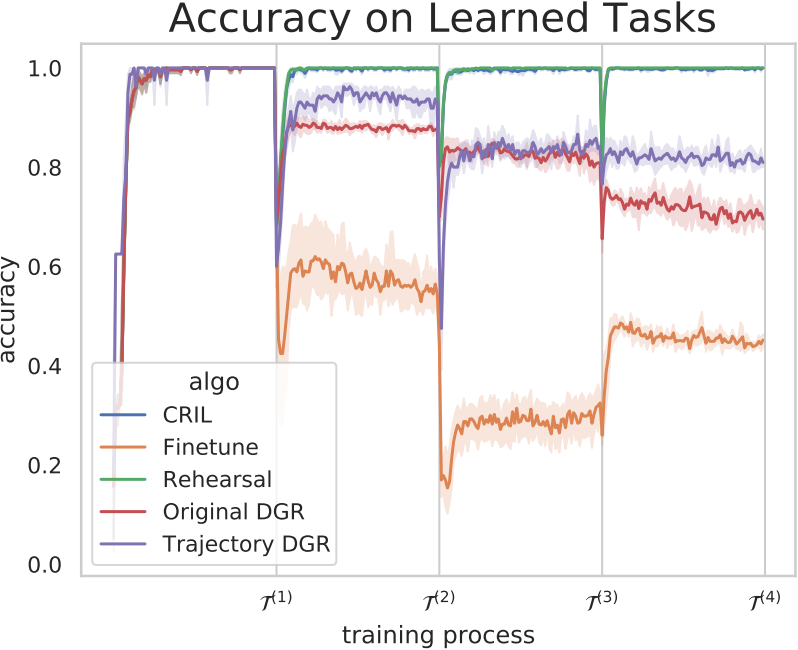}
		\end{minipage}
	}

	\caption{Accuracy of all tasks learned so far in the training process of different methods in both simulation and real world environments. Each method is tested using 5 random seeds.}
	\label{fig:res}
	
\end{figure}

Table \uppercase\expandafter{\romannumeral1} and Fig. \ref{fig:res} show the quantitative results and the forgetting curves of different methods. Since Finetune does not carry out any kind of continual learning approach, its performance abruptly degrades across the continual learning process. Original DGR and Trajectory DGR can maintain the accuracy on the first few tasks, but their performances also drop when the number of tasks increases. Our proposed method achieves promising performances that are close to the upper bound performance of rehearsal in both simulation and real-world environments.

\begin{figure}[t]
	\centering
	\subfigure[Simulation Results]{
		\begin{minipage}[t]{0.46\linewidth}
			\centering
			\includegraphics[width=1\linewidth]{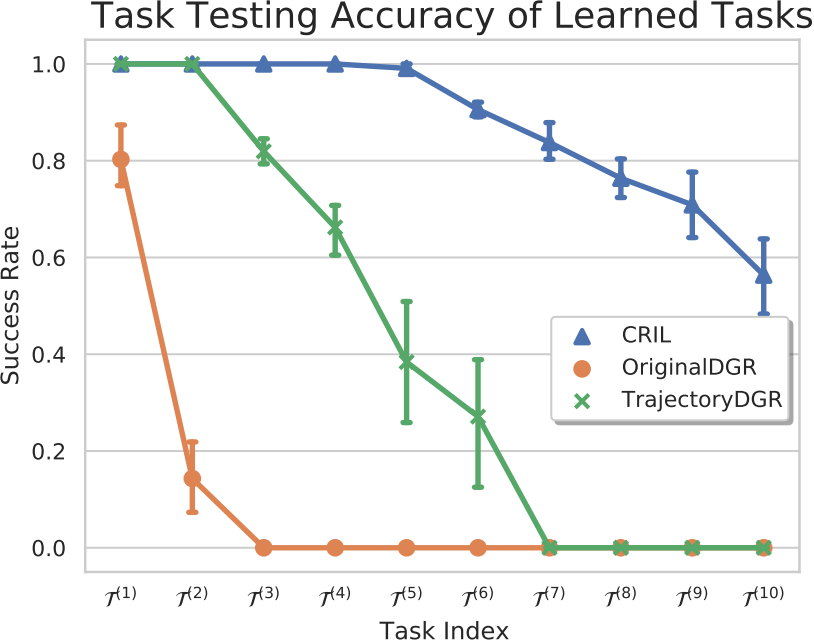}
		\end{minipage}
	}
	\subfigure[Real world Results]{
		\begin{minipage}[t]{0.46\linewidth}
			\centering
			\includegraphics[width=1\linewidth]{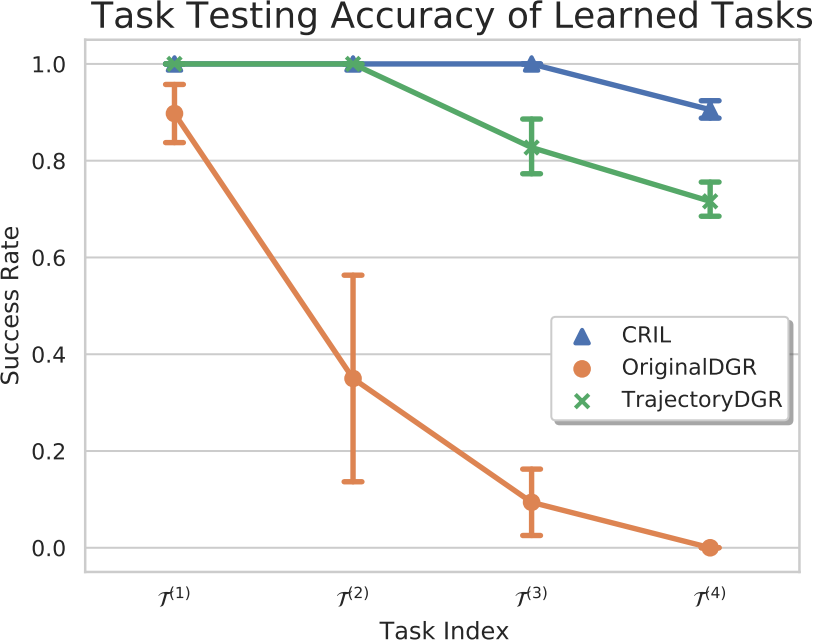}
		\end{minipage}
	}
	
	\caption{Success rates of all learned tasks of different methods in testing phase with means and standard deviations from 5 test trials.}
	\label{fig:test}
	\vspace{-0.2cm}
\end{figure}

As robot IL is not an image classification problem but
a sequential decision-making problem, in which the robot
must be placed in environments to accomplish tasks
by reproducing expert demonstrations, the task success rate
should be considered as another important performance metric. Fig. \ref{fig:test} shows the success rates of three methods when they are employed to operate in tasks, where CRIL outperforms
conventional DGR methods by a large margin. In this situation, the statistical accuracy of state-action mappings becomes less important, since even if the robot makes mistakes in only one step, it will probably fail to complete the whole task. Note these results are also correlated to the IL algorithm employed for continual learning. For instance, using IRL or GAIL may lead to better results.

It is worth mentioning that the training time of Original DGR and Trajectory DGR are both much longer than CRIL, since the generators in the first two methods are hard to train. For Original DGR, it becomes more serious since the generated images are separated and cannot construct complete trajectories. Thus in order to cover states in a trajectory as much as possible, in our experiments, we generate 10 times more samples in Original DGR method than the others, and this leads to a slower training process.

\subsection{Qualitative Analysis of Generated Images}

Fig. \ref{fig:generatedimages} shows the generated trajectories of three different methods. CRIL produces out the highest quality images. For Original DGR, the generated images are independent of each other, thus cannot form a complete trajectory. It is interesting that, in the generated images of CRIL, the first frames of each trajectory are also in poor qualities (see the results in simulation environments in Fig. \ref{fig:generatedimages}). However, subsequent frames in these trajectories can go back to high qualities, which reveals the power of our prediction model. Since the prediction model is learned by supervised learning, its output can easily and always be a high quality image compared to the generative model. This makes the whole model have strong robustness, thus most parts of generated trajectories are high quality images. This situation also shows that even if we have limited the generator to producing only the first frames, generating high quality images is still very hard for a GAN network as the number of tasks increases.

\section{CONCLUSION AND FUTURE WORK}

In this work, we present a novel method to apply DGR  in robot IL tasks to achieve the continuous learning ability. The main idea is to decompose the full trajectory generation problem to a
first-frame generation problem and an action-conditioned
next-frame prediction problem. For future work, researchers might consider how to apply DGR with multi-modal demonstrations, which may contain conflicting actions on the same states, which brings challenges for the predictor. Another attractive direction would be studying how to stably generate pseudo data in the DGR process, such as learning an  embedding space rather than directly generate high-dimensional raw images.

\section*{ACKNOWLEDGMENT}

We would like to thank Xin Su, Zhile Yang and Yizhou Jiang for various discussions. This work was supported in part by the National Natural Science Foundation of China under Grant 61671266 and Grant 61836004, in part by the Tsinghua-Guoqiang research program under Grant 2019GQG0006, and in part by Qualcomm Technologies, Inc.

\bibliographystyle{IEEEtran.bst}
\bibliography{IEEEabrv,IEEEexample.bib}

\end{document}